# Multi-Agent Motion Planning using Deep Learning for Space Applications

Kyongsik Yun[1], Changrak Choi[2], Ryan Alimo[3], Anthony Davis[4], Linda Forster[5], and Amir Rahmani[6]

*NASA Jet Propulsion Laboratory, California Institute of Technology, Pasadena, CA, 91109, USA*

Muhammad Adil[7] and Ramtin Madani[8]

*University of Texas at Arlington, Arlington, TX, 76019, USA*

**State-of-the-art motion planners cannot scale to a large number of systems. Motion planning for multiple agents is an NP (non-deterministic polynomial-time) hard problem, so the computation time increases exponentially with each addition of agents. This computational demand is a major stumbling block to the motion planner's application to future NASA missions involving the swarm of space vehicles. We applied a deep neural network to transform computationally demanding mathematical motion planning problems into deep learning-based numerical problems. We showed optimal motion trajectories can be accurately replicated using deep learning-based numerical models in several 2D and 3D systems with multiple agents. The deep learning-based numerical model demonstrates superior computational efficiency with plans generated 1000 times faster than the mathematical model counterpart.**

## I. Nomenclature

| | | |
|---|---|---|
| $X(t)$ | = | state of agent |
| $O(t)$ | = | state of obstacle |
| $U(t)$ | = | control input |
| $n$ | = | mean motion of chief spacecraft |
| $T$ | = | time allowed to reach goal state |
| $E$ | = | mean squared error |
| $w$ | = | weight |
| $b$ | = | bias |

## II. Introduction

Motion planning solves the problem of how best to move the system from its current state to a given target state while meeting constraints such as collision avoidance [1]. This is essentially a computationally complex NP-hard problem, and the single agent motion planning problem has proven to be PSPACE-complete (the hardest problem in polynomial space) [2]. Complexity increases due to a number of factors, including the geometry of the robot and obstacles, the dynamics of the robot, the objective cost and the estimated uncertainty. Several planning approaches exist in the literature that addressed various aspects of this complexity. Sampling-based methods with the probabilistic roadmap (PRM) [3] and the rapidly-exploring random tree (RRT) [4,5] variants are widely used because of their practicality and their ability to scale up to many degrees of freedom. The graph-based method via discretization with A* variants [6] is preferred because of its ability to efficiently generate optimal paths for low-dimensional problems. Optimization-based methods such as Sequential Convex Programming (SCP) [7,8] can handle objective cost and system dynamics well, but suffer from the high non-convexity of the problem.

---

[1] Technologist, Autonomous Systems Division
[2] Robotics Technologist, Autonomous Systems Division
[3] Data Scientist, Mission Systems and Operations Division
[4] Scientist, Science Division
[5] JPL Postdoctoral Fellow, Science Division
[6] Group Supervisor, Autonomous Systems Division
[7] PhD Student, Department of Electrical Engineering
[8] Assistant Professor, Department of Electrical Engineering

The motion planning problem becomes significantly harder when scaled from a single agent to multiple agents [9]. The computational complexity is PSPACE-hard even for simple robot geometry [10], and the complexity becomes NEXPTIME when dynamic parameters are involved [11]. In particular, when the number of agents increases and becomes a swarm, these swarm motion planning problems cannot be solved with the state-of-the-art planning methods mentioned above. Sampling-based methods [12–14] often fail to converge on a solution in a reasonable amount of time. The discrete graph-based method can scale to hundreds [15,16] through path generation, but it cannot handle dynamic systems that have drifts or are non-holonomic. The optimization-based method can handle dynamic systems, but has low computational processing capability for large multi-agent systems [17].

In the aerospace industry, the demand for future missions involving swarms of space vehicles continues to grow in both the public and private sectors [18]. These missions cannot rely solely on traditional ground instructions for operations, as the number of spacecraft involved becomes so large. The autonomy of the spacecraft is not only desirable but also necessary for the operation of the swarm system, and motion planning for autonomy is one of the key components. However, the lack of scalability of the existing multi-agent motion planning approach greatly limits its applicability to swarm missions.

In this study, we tackled the computational complexity of a multi-agent motion planning problem using deep learning. We transformed a computationally demanding mathematical motion planning problem into a deep learning-based numerical problem. A deep neural network architecture was constructed that can handle multiple agents (~100) and their interactions with other agents and obstacles.

A recent study showed that a deep neural network can accurately learn a nonlinear tracking controller behavior in multirotor swarms [19]. Another recent study showed learning decentralized controllers for robot swarms with graph neural networks [20]. These previous studies successfully showed accurate modeling of nonlinear behavior of multiple agents using deep learning. The next step would be to investigate path planning of swarms using deep neural networks.

In this study, deep neural networks were applied to the swarm path planning. The aim of this study was to improve the computational efficiency of existing mathematical optimization modeling, while maintaining the accuracy of swarm path planning trajectories. Previously, researchers showed that modeling accuracy was improved by combining deep neural networks and swarm optimization algorithms [21,22]. Spiking neural networks were also used to model swarm optimization solvers [23]. These authors demonstrated the efficient implementation of such neural dynamics in a neuromorphic hardware system, showing the potential of high performance combined with high energy efficiency [23]. Only recently complex swarm dynamics modeling based on machine learning or deep learning started to be investigated compared to mathematical optimization models. We investigated multi-agent path planning in 2D and 3D orbital transfer dynamics. The performance and computational efficiency of a deep learning model was compared with a conventional convex optimization model.

### III. Problem Definition and Dataset

We investigated a deep learning-based method for solving swarm motion planning for two major use cases of space applications. The first is addressing spacecraft in a gravity-free environment. This relates to a mission scenario where a fleet of spacecraft in deep space needs to be reconfigured into different shapes on-demand to achieve mission objectives. A similar situation can be observed if a team of free-flying robots (e.g. Astrobee) operating on a larger spacecraft (e.g. the International Space Station) must move efficiently from one section to another. The second is addressing spacecraft in low Earth orbit where gravity affects motion. This is relevant when a large number of deputy spacecraft (e.g. CubeSats) need to change their orbit and position relative to the chief spacecraft to perform in-orbit inspection or docking for assembly. Both of these cases represent a common situation that can arise in missions involving a spacecraft swarm, where multiple spacecraft must actively maneuver in the desired configuration to perform the required task.

Each of the above two cases has motion dynamics. The spacecraft motion of the former is controlled by the double integrator dynamics, where the thrust directly produces a proportional acceleration in that direction. In the latter, the equation of motion involves the effect of gravity, and can be linearized to the Clohessy-Wiltshire equation [24] when described in relation to the main spacecraft in a circular orbit. In both cases, the goal of the swarm motion planning is to generate a minimum fuel trajectory for each spacecraft in the swarm, transitioning from the current state to the assigned goal state. More specifically, the objective function is to minimize the total L1-norm of fuel consumption for the entire team of spacecraft. Each spacecraft is considered homogenous of the same shape and has full control over its operation, which can apply thrust in all x, y, z directions. The trajectory generated must be within given actuation limits in each direction, and collisions involving given obstacles and other spacecraft in the environment must be avoided.

We created datasets with increasing complexity from 1 to 10 agents to solve single and multi-agent problems. Each data set contains the optimal trajectories (ground truth) for a defined problem that is required to fully train a deep neural network. In order to generate the optimal motion trajectory, we used an optimization-based method that utilizes both sequential convex programming and recently developed parabolic relaxation [25]. The problem definition is first transformed into the form of the quadratic constrained quadratic programming (QCQP) with the corresponding objective function and dynamical/physical constraints. The QCQP formulation is then processed through a series of transformations including relaxation and penalization to efficiently generate optimal trajectories.

### A. Gravity-Free 2D Double Integrator (1 Agent and 1 Obstacle)

We first tackled the simple problem of a single agent moving in a 2D environment without gravity. The agent must move from the initial position to the target position while avoiding any single static obstacles present. The transition must be done within the T time step, minimizing the control input (fuel consumption). The state of the agent and obstacle is a 4 degrees of freedom (dof) parameter that includes both position and velocity. The control input is a 2 dof parameter in both $x$ and $y$ directions and the motion is controlled by 2D double integrator dynamics.

$$\text{State of Agent: } X(t) = \begin{bmatrix} x(t) \\ y(t) \\ \dot{x}(t) \\ \dot{y}(t) \end{bmatrix}, \quad \text{State of Obstacle: } O(t) = \begin{bmatrix} a(t) \\ b(t) \\ \dot{a}(t) \\ \dot{b}(t) \end{bmatrix} \quad (1)$$

$$\text{Control Input: } U(t) = \begin{bmatrix} u_1(t) \\ u_2(t) \end{bmatrix} \quad (2)$$

$$\text{Dynamics (2D Double-Integrator): } \ddot{x}(t) = u_1(t), \ \ddot{y}(t) = u_2(t) \quad (3)$$

The shape of the agent is assumed to be circular with a radius of 0.1. Although the actual shape of the spacecraft is different, using an outer circular approximation of the agent shape is a standard practice in motion planning that simplifies collision detection. The radius of the obstacle is set to 0.1 and an additional minimum clearance distance of 0.1 is applied between the spacecraft and the obstacle, acting as a safety buffer to account for possible conditions and operational uncertainties.

The data set has the following inputs: First, a single agent starts at position (0,0) in a stationary state (speed 0) and reaches a position (1,1) where it stops at all problem instances. The position of the obstacle (a, b) is chosen randomly between 0 and 1 in a uniform distribution. The control input (thrust in x and y directions) is limited to a maximum of 0.1, and the number of time steps to reach the target position is set to T = 10.

$$X(t=0) = \begin{bmatrix} 0 & 0 & 0 & 0 \end{bmatrix}^T, \quad X(t=T) = \begin{bmatrix} 1 & 1 & 0 & 0 \end{bmatrix}^T \quad (4)$$

$$O(t) = \begin{bmatrix} a_{rand(0,1)} & b_{rand(0,1)} & 0 & 0 \end{bmatrix}^T \quad (5)$$

$$|u_1(t)| \leq 0.1, \quad |u_2(t)| \leq 0.1 \quad (6)$$

The data set has output in the form of trajectories and control inputs. The trajectory gives the state of the agent $X(t)$ at each time step from t = 0 to t = T and the control input specifies the action $U(t)$ applied at each time step from t = 0 to t = T-1, with T = 10.

$$X = \begin{bmatrix} 0 & x(1) & . & . & x(T-1) & 1 \\ 0 & y(1) & . & . & y(T-1) & 1 \\ 0 & \dot{x}(1) & . & . & \dot{x}(T-1) & 0 \\ 0 & \dot{y}(1) & . & . & \dot{y}(T-1) & 0 \end{bmatrix}, \quad U = \begin{bmatrix} u_1(0) & . & . & u_1(T-1) \\ u_2(0) & . & . & u_2(T-1) \end{bmatrix} \quad (7)$$

**B. Gravity-Free 2D Double Integrator (10 Agents)**

This issue is defined in the same way as the single agent case above, only the differences are detailed here. First, the number of agents increases from 1 to 10, so now the state is a 40 dof parameter that describes the position and velocity of all 10 agents. The control input is also a 20 dof parameter that specifies the behavior of all 10 agents. In this multi-agent setup, each agent now acts as an obstacle that other agents should avoid. Because of this characteristic, the single obstacle required for collision avoidance training is no longer required and thus removed from the problem definition. The 10 agents are initially randomly placed inside a square of unit dimension. The goal state is defined so that 10 agents form a circular shape with a radius of 0.6 at the center of the unit square. Each agent has a radius of 0.05 and an additional minimum spacing of 0.05 is imposed.

**C. Relative Orbit Transfer near Earth (1 Agent)**

In this problem, the deputy spacecraft is in orbit around the Earth, and its motion is explained in relation to the chief spacecraft in a circular orbit. We assume that the relative distance between the deputy and the chief is sufficiently close and ignore the J2 perturbation effects such that the Clohessy-Wiltshire equation of motion holds. The deputy's state is described in the Local-Vertical, Local-Horizontal (LVLH) frame attached to the chief, where the x direction points radially away from Earth towards the chief, the y direction points the direction of velocity of the chief, and the z direction points to the direction of the chief's angular momentum. The state is a 6 dof parameter that includes both relative position and velocity. The control input is a 3 dof parameter that describes the thrust in the x, y and z directions, respectively.

State of Agent: $X(t) = \begin{bmatrix} x(t) \\ y(t) \\ z(t) \\ \dot{x}(t) \\ \dot{y}(t) \\ \dot{z}(t) \end{bmatrix}$, Control Input: $U(t) = \begin{bmatrix} u_1(t) \\ u_2(t) \\ u_3(t) \end{bmatrix}$ (8)

Dynamics (CWH):
$$\ddot{x}(t) = 3n^2 x(t) + 2n\dot{y}(t) + u_1(t)$$
$$\ddot{y}(t) = -2n\dot{x}(t) + u_2(t)$$
$$\ddot{z}(t) = -n^2 z(t) + u_3(t)$$ (9)

Also, we assume that the deputy spacecraft is in a stable passive relative orbit (PRO) centered around the chief, which is a thrust-free orbit that can maintain a certain relative distance from the chief spacecraft. Concentric PRO ensures collision avoidance between deputies [26]. This imposes an energy matching condition of Eq 10 and a concentric PRO condition of Eq 11. Similar to the case without gravity, we used a solid outer sphere approximation to represent the geometry of the deputy. The radius of the deputy is set to 5 and an additional minimum clearance of 5 is imposed for safety.

Energy matching: $\dot{y}(0) = -2nx(0)$ (10)

Concentric PRO in X-Y plane about origin: $\dot{x}(0) = \frac{1}{2} ny(0)$ (11)

The data set has the following inputs: The deputy spacecraft is initially in a PRO with a phase given at time t = 0 and must transfer to another PRO with a phase specified at time t = T. The starting and target PROs are chosen in such a way that they are constrained to the x-y plane for simplicity and to the semi-major axis randomly selected between 25 and 75 in a uniform distribution. The initial and target phases within the PRO are also randomly chosen between 0 and $2\pi$. The control input is limited to a maximum of 10 thrust forces in each x, y, z direction. The number of time steps to reach the target state is set to T = 100.

The data set has output in the form of trajectories and control inputs. The trajectory gives the state of the agent X(t) at each time step from t = 0 to t = T and the control input specifies the actuation U(t) applied at each time step from t = 0 to t = T-1, with T = 10.

$$X = \begin{bmatrix} x(0) & . & . & x(T) \\ y(0) & . & . & y(T) \\ z(0) & . & . & z(T) \\ \dot{x}(0) & . & . & \dot{x}(T) \\ \dot{y}(0) & . & . & \dot{y}(T) \\ \dot{z}(0) & . & . & \dot{z}(T) \end{bmatrix}, \quad U = \begin{bmatrix} u_1(0) & . & . & u_1(T-1) \\ u_2(0) & . & . & u_2(T-1) \\ u_3(0) & . & . & u_3(T-1) \end{bmatrix}$$ (12)

### D. Relative Orbit Transfer near Earth (10 Agents)

This issue is a direct extension of the single agent case above. Instead of having only one deputy spacecraft, now 10 deputy spacecraft perform PRO transfers simultaneously. The problem definition is the same as above, but the state and control inputs are different. The state is now a 60 dof parameter describing the position and velocity of all 10 deputy spacecraft, and similarly the control input is a 40 dof parameter. Fig 1 shows an example of a problem instance defined here for a single deputy and all 10 deputy cases.

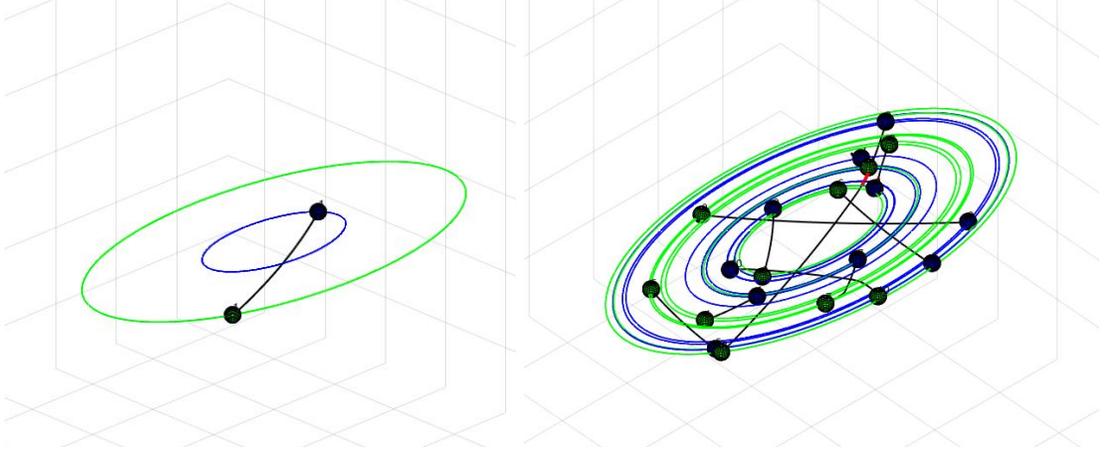

**Fig 1.** Examples of passive relative orbit (PRO) transfers for a single deputy spacecraft (left) and 10 deputy spacecraft (right). The deputy must be transferred from the initial PRO to the target PRO where the phases are assigned. The minimum fuel consumption trajectory is computed over multiple instances of the problem to work with the dataset required for deep neural network training.

## IV. Deep Neural Network Architecture

We used a multilayer neural network made up of multiple densely connected layers with an activation function and a dropout layer. We used rectified linear units for the nonlinear activation function: $g(x) = max(0, x)$ where $x$ is the input to a neuron. Rectifier is the most used activation function for deep neural networks that have biological validity and better gradient propagation with less vanishing gradient problems. We used a dropout regularization rate of 0.5 to prevent overfitting.

Neural network training aims to minimize the error function $E$, the mean squared error (MSE) that quantifies the difference between the computed output trajectory and velocity of the neural network, and the true trajectory and velocity of the mathematical model $\{X, Y, X', Y'\}$ for input initial and final position and velocity. Given N time steps, $\{X, Y, X', Y'\}$ is defined as $\{(x_1, y_1, x_1', y_1'), ..., (x_N, y_N, x_N', y_N')\}$. The error function $E$ is

$$E(\widehat{X}, \widehat{Y}, \widehat{X}', \widehat{Y}') = \frac{1}{2N} \sum_{i=1}^{N}((\widehat{x}_i - x_i)^2 + (\widehat{y}_i - y_i)^2 + (\widehat{x}_i' - x_i')^2 + (\widehat{y}_i' - y_i')^2)$$

$$= \frac{1}{2N} \sum_{i=1}^{N} \sum_{j=1}^{2} ((g(w_{x_j} \cdot x_j + b_{x_j}) - x_i)^2 + ((g(w_{y_j} \cdot y_j + b_{y_j}) - y_i)^2 +$$

$$((g(w_{x_j'} \cdot x_j' + b_{x_j'}) - x_i')^2 + ((g(w_{y_j'} \cdot y_j' + b_{y_j'}) - y_i')^2) \quad (13)$$

where $\{\widehat{x}_i, \widehat{y}_i, \widehat{x}_i', \widehat{y}_i'\}$ denotes the output of the neural network on input $\{x_j, y_j, x_j', y_j'\}$ where $j = \{1, 2\}$ for the start and stop position and velocity. The objective is to change $w$ and $b$ such that $E$ is as close to zero as possible. The weights and biases were updated according to the equation

$$w_{i+1} = w_i - \alpha \frac{\partial E(X, Y, \widehat{X}, \widehat{Y})}{\partial w_i} \quad (14)$$

$$b_{i+1} = b_i - \alpha \frac{\partial E(X, Y, \widehat{X}, \widehat{Y})}{\partial b_i} \tag{15}$$

where $w_i$ and $b_i$ are the values of $w$ and $b$ after the $i$ th iteration of gradient descent, and $\frac{\partial f}{\partial x}$ is the partial derivative of $f$ with respect to $x$. $\alpha$ is the learning rate where $\alpha=0.001$. Based on the chain rule, the weight delta and bias delta are

$$\Delta w = w_{i+1} - w_i = \frac{1}{N} \sum_{i=1}^{N} \sum_{j=1}^{2} \alpha(x_i - \widehat{x}_i) g'(h_i) x_j \tag{16}$$

$$\Delta b = b_{i+1} - b_i = \frac{1}{N} \sum_{i=1}^{N} \sum_{j=1}^{2} \alpha(x_i - \widehat{x}_i) g'(h_i) \tag{17}$$

where $\widehat{x}_i = g(w \cdot x_j + b)$ and $h_i = w \cdot x_j + b$. After the completion of training, the weights and biases are optimized to minimize the error function $E$.

The number of layers and parameters were empirically determined by investigating the performance of several different layers (3-12) and parameters (10-200). We found that 100 parameters per layer and 4 layers showed the minimum mean squared error (MSE) (Fig 2).

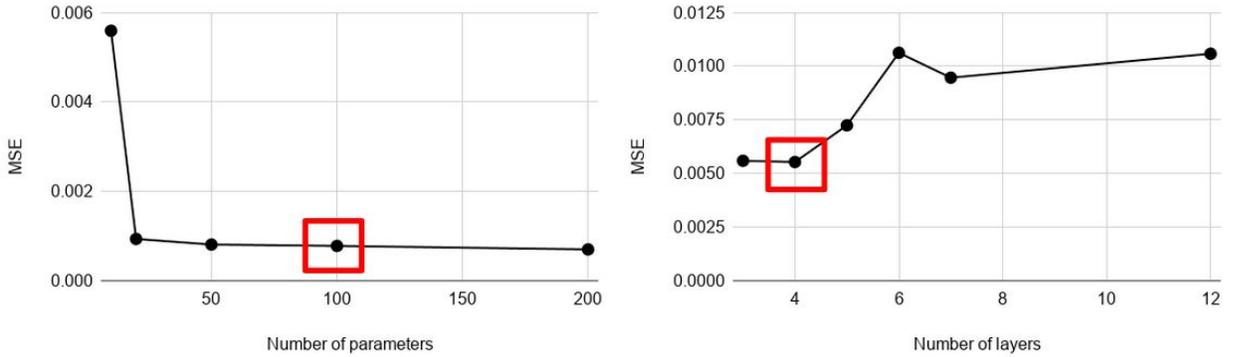

**Fig 2. Parameter optimization by examining different numbers of parameters and layers of neural networks. MSE is minimized in 100 parameters and 4 layers.**

A. **2D Double Integrator**

The simplest toy model, a 2D double integrator, a second-order control system was used to model the dynamics of a simple agent in a two-dimensional space under the influence of a time-varying force input. For each agent, 12 parameters were used as input, including initial position $(x_i, y_i)$, velocity $(x_i', y_i')$, acceleration $(x_i'', y_i'')$, and final position $(x_f, y_f)$, velocity $(x_f', y_f')$, acceleration $(x_f'', y_f'')$. For each obstacle, 3 parameters were used as input, including position $(x_o, y_o)$ and radius $(r)$. Considering 10 agents, the total number of input parameters is 120.

The output size of the network depends on the temporal resolution. For the number of time steps, T = 10, 60 parameters are determined as the output (T = 10 time steps × 6 (position, velocity, acceleration) = 60 parameters). Among the output parameters, only the trajectory and velocity were included, and the acceleration information can be excluded (only 40 output parameters left). This is because the acceleration information can be computed later using the differential relationship between acceleration, velocity, and position. Reducing the number of output parameters affects the number of optimized hidden layers and the required parameters, so we used this method to save memory and increase computational efficiency.

B. **3D Passive Relative Orbit (PRO) Transfer**

As a more realistic model, we investigated PRO transfer. For each agent, 18 parameters were used as input, including initial position $(x_i, y_i, z_i)$, velocity $(x_i', y_i', z_i')$, acceleration $(x_i'', y_i'', z_i'')$, and final position $(x_f, y_f, z_f)$, velocity $(x_f', y_f', z_f')$, acceleration $(x_f'', y_f'', z_f'')$. Considering 10 agents, the total number of input parameters is 180. Similarly, the output size of the network was 90 parameters (T = 10 with 9 position, velocity, and acceleration parameters). Compared to the 2D double integrator, 3D PRO transfer differs in that the path planning requires not only additional dimensions but also gravity constraints to be considered.

## V. Deep Neural Network Training and Testing Results

### A. 2D Double Integrator

A single agent and a single obstacle case were tested as the simplest model. The results showed accurate position and velocity estimation (RMSE = 0.0129 ± 0.0088) using a deep learning-based numerical model (Fig 3). The computational efficiency per estimated condition improved from a few seconds to $5 \times 10^{-4}$ seconds (approximately 1000 times improvement). Neural network training took from 3 to 5 hours (NVIDIA GTX 1080).

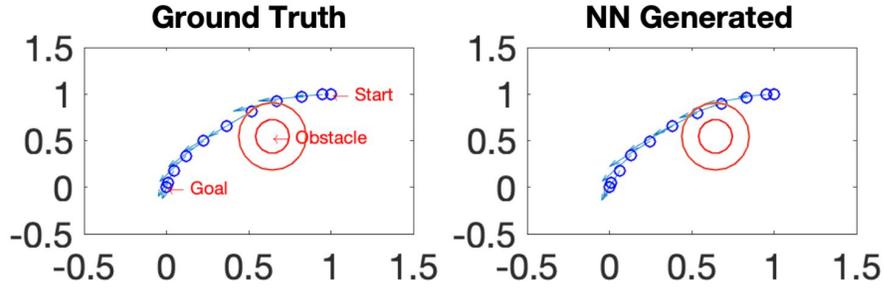

**Fig 3. An example of accurate location, velocity and acceleration estimation of a neural network numerical model compared to the ground truth mathematical model (RMSE =0.0108). Blue circles represent single agent trajectories. Arrows indicate velocity at a given position. A single obstacle marked in red is fixed (radius = 0.1). The outer red circle of the obstacle represents the safety distance to avoid collision between the obstacle and the agent. It is defined as the safety radius = obstacle radius + 0.1.**

The next step was to investigate a 2D 10 agents model with various training data sizes. Deep neural networks were trained using 1,000, 10,000, and 100,000 data. We randomly selected 70% of the given trajectory data for training, 15% of the data for validation during training, and 15% of the data for final testing. Training stopped when validation loss did not decrease to prevent overfitting. As more training data were added, the trajectory generated by neural networks became closer to that of the ground truth (Fig 4).

We added the acceleration of each agent at each time step using the L1 norm, which represents the fuel consumption of each agent as a measure of model performance. As the number of training data increased, the fuel consumption of neural network-based trajectory estimation decreased. When training with 100,000 data, fuel consumption fell close to the ground truth fuel consumption (mean ground truth fuel consumption = 1.91±0.46, mean neural net fuel consumption = 2.23±0.21). The difference between the two groups was not significant (P=0.24, two-sample t-test) (Fig 5).

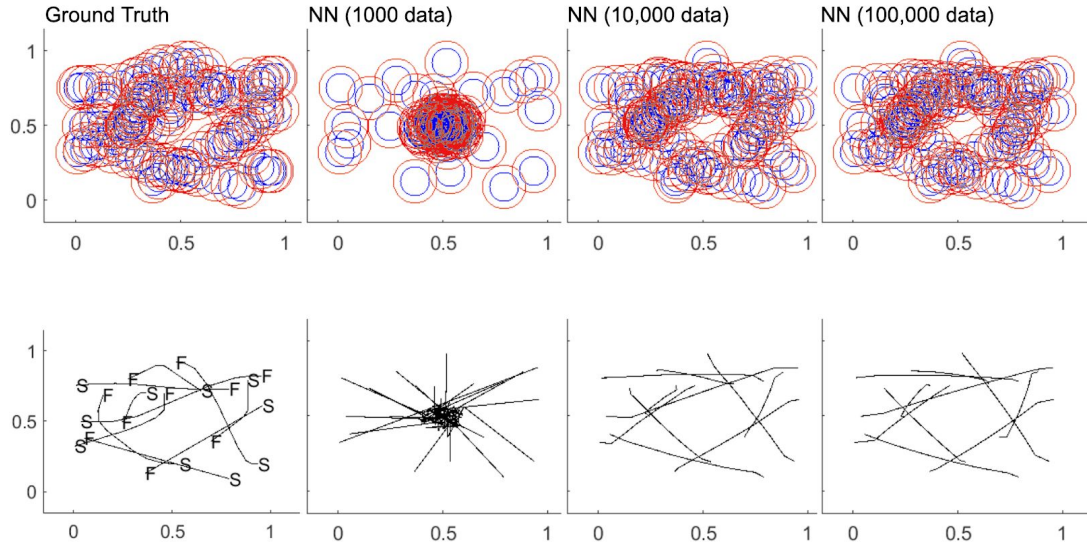

**Fig 4. An example of a 2D 10 agent model with various training data sizes. Trajectory modeling improved as the number of training data increased. Blue circles represent agent trajectories. Red circles represent the safety distance to avoid collision between the agents. In the bottom left plot, text "S" represents the start position, and text "F" represents the final position.**

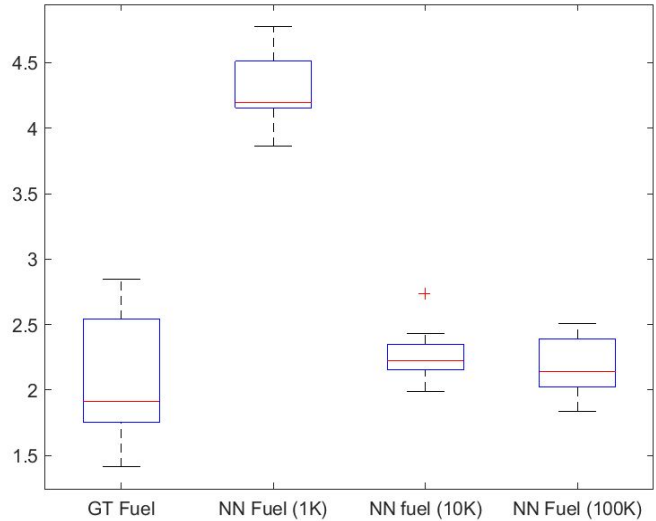

**Fig 5. Fuel consumption of the ground truth and neural network models with different training data sizes. The fuel consumption of neural network modeling decreased as the number of training data increased.**

B. **3D PRO Transfer**

We also tested neural network-based PRO transfer estimation in 3D space with gravity constraints. We found that the path was accurately estimated using neural networks (Fig 6). It took 0.99 seconds to generate 2,000 path estimation results using a neural network and 1.7 hours to generate the same 2,000 optimal path plans using the convex optimization model, showing a 6,000-fold increase in computational efficiency.

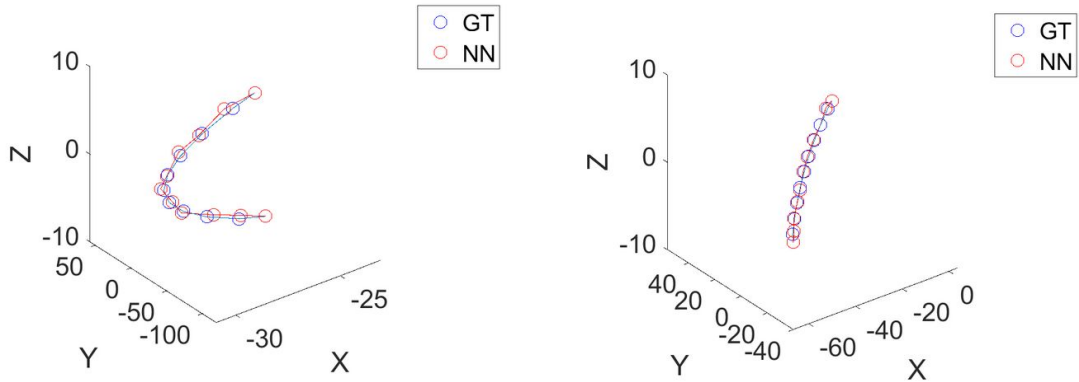

**Fig 6. Representative examples of accurate path estimation of a single agent model in 3D PRO transfer conditions. The path generated by the neural network was very similar to the path of the ground truth. The computational efficiency improved by 6,000 times using neural networks.**

We expanded the problem with 10 agents in 3D space. The problem was much more complex. We trained the neural network using 30,000 ground truth data, but the neural network did not succeed in learning collision avoidance and path planning of 10 agent interactions (Fig 7). As a result, the neural network fuel consumption was much higher than the optimal ground truth fuel consumption. We assume that the neural network training was largely unsuccessful because of the high complexity of the problem (10 agents in 3D space) and insufficient training data size. We used 30,000 training data for this result, and we suggest generating up to 100,000 training data to provide the full breadth of 3D 10 agent dynamics for training neural networks. Another solution would be to limit the complexity of the problem by limiting parameters such as start and stop locations of each agent to the same location in all datasets.

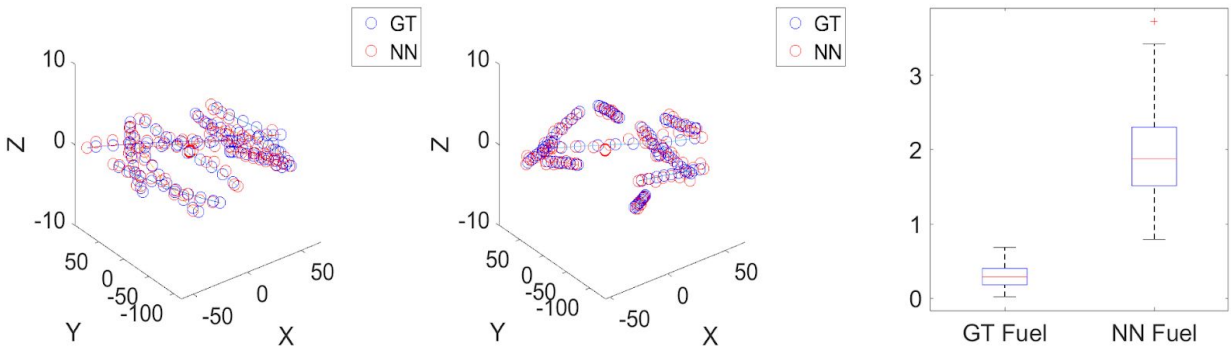

**Fig 7. Inaccurate path estimation examples of neural networks in a 10 agent 3D model. The trajectory of 10 agents did not follow the ground truth, and the fuel consumption of the neural network model was much higher than the ground truth.**

## VI. Application to Space

The results of this study demonstrate the potential that deep neural network planner has in generating minimum fuel optimal trajectories with training, at a significantly low computational cost compared to traditional trajectory optimization. Although the technology is at its infancy and has to be matured to be flight-ready, it opens up the possibility of having real-time and on-board computation of trajectories for a mission that involves a large fleet of spacecraft. Such capability will act as a key component of autonomy that enables operation of future swarm spacecraft missions that hold many opportunities for the space industry in general [18].

The first category of swarm spacecraft missions that can greatly benefit from is on-orbit servicing. With advent of SmallSats and ever increasing number of SmallSats being launched by both the public and private space industry (e.g. Starlink from SpaceX), there is growing need for servicing of space assets in orbit around the Earth [27]. The on-orbit servicing may consist of inspecting, docking, and manipulating a space asset, all of which could be greatly enhanced by use of multiple spacecraft performing the task [28]. On-orbit servicing using a fleet of spacecraft has to have a real-time trajectory planning to avoid collisions and perform necessary reconfiguration for a given task. With a deep neural network planner on-board, active reconfiguration of a large fleet of spacecraft can be performed even with constrained on-board computation power.

The second category is Earth science missions that benefit from having multiple measurements of Earth phenomena. For example, multiple measurements from hundreds of SmallSats in low Earth orbit can be used to gather the information needed to generate a high-resolution spatiotemporal 3D cloud map. Multi-angle observations already allow estimation of the cloud top height and horizontal position by stereo-imaging and tracking of image features [29]. A tomographic approach yields volumetric information about clouds [30,31]. Specifically, the 3D distribution of liquid water combined with the droplet size distribution are essential to improve the representation of clouds in climate models and cloud process modeling.

Furthermore, the benefit of multi-angle observations from a low Earth orbit is not only constrained to clouds, but provides valuable information about aerosol plumes: 3D information about the mass and particle properties of smoke and dust plumes helps advance monitoring and nowcasting of air quality and the spread of wildfires, volcanoes, and sand storms.

## VII.  Conclusion

In this study, we showed that deep neural networks can accurately estimate multi-agent path plans using training data generated from existing mathematical models. Models were tested with single and 10 agents in 2D and 3D space. The accuracy of the generated trajectories and the fuel consumption were comparable to those of the ground truth. It also showed that the computational efficiency dramatically improved (1,000x-6,000x depending on the complexity of the problem), which can be applied to enable real-time, distributed, on-board multi-agent path planning for large fleets of spacecraft (>100). The results of this study demonstrate the potential of deep neural network models in bridging the technology gap that exists in the scalability of multi-agent motion planning. This technical capability is essential for operating large-scale spacecraft that cannot rely solely on ground guidance and enables future scientific missions that will greatly benefit from multi-angle space measurements. For example, multiple measurements from hundreds of SmallSats in low Earth orbit can be used to gather the information needed to generate a high-resolution spatiotemporal 3D cloud map, which solves the major uncertainties in climate science and the challenges of weather forecasting.

In order to mature the deep neural network-based planning technology, it is necessary to increase the path planning accuracy of the neural network. Therefore, we propose a hybrid approach that utilizes a convex optimization model to fine tune the trajectories generated by the neural network to improve accuracy and robustness. Moreover, reinforcement learning is considered as an unsupervised, self-learning, adaptive system independent of the convex optimization model.

## Acknowledgments

This research was carried out at the Jet Propulsion Laboratory, California Institute of Technology, under a contract with the National Aeronautics and Space Administration (80NM0018D0004).